# How to Measure Gender Bias in Machine Translation: Optimal Translators, Multiple Reference Points

**Anna Farkas\*, Renáta Németh\*\***

\* ELTE Eötvös Loránd University, Faculty of Social Sciences, Budapest

\*\* Research Center for Computational Social Sciences, ELTE Eötvös Loránd University, Faculty of Social Sciences, Budapest,

## Abstract

In this paper—as a case study—we present a systematic study of gender bias in machine translation with Google Translate. We translated sentences containing names of occupations from Hungarian, a language with gender-neutral pronouns, into English. Our aim was to present a fair measure for bias by comparing the translations to an optimal non-biased translator. When assessing bias, we used the following reference points: (1) the distribution of men and women among occupations in both the source and the target language countries, as well as (2) the results of a Hungarian survey that examined if certain jobs are generally perceived as feminine or masculine. We also studied how expanding sentences with adjectives referring to occupations effect the gender of the translated pronouns.

As a result, we found bias against both genders, but biased results against women are much more frequent. Translations are closer to our perception of occupations than to objective occupational statistics. Finally, occupations have a greater effect on translation than adjectives.

## Introduction

In recent years, there has been a growing interest in the research of machine bias, also referred to as algorithmic bias. The term "machine bias" describes the phenomenon that machine learning algorithms are prone to reinforce or amplify human biases (Prates et al., 2020). Machine learning algorithms are written by humans and draw conclusions from data that was created, collected, cleaned, and stored by humans. Thus, human error and bias cannot be eliminated from the operation of algorithms and from the results they generate (Ságvári, 2017). Nowadays, machine learning is used in a wide variety of sectors, including insurance, crime prevention, recruitment, healthcare, search engines, news outlets, online advertising, and recommendation systems among others (Burrell, 2016; Goodman and Flaxman, 2016; Sandvig et al., 2014; Ságvári, 2017). Since machine learning algorithms have a great deal of influence on many aspects of life, it raises concerns when those algorithms exhibit bias or discrimination. Over the past few years, researchers and journalists have discovered many cases when algorithms created biased results against certain social groups, thus, making socially unjust decisions in terms of race, gender, age, or religion—a few examples of these are: gender bias in hiring algorithms (Chen et al., 2018; Dastin, 2018; Schwarm, 2018), ageist and racist ad targeting (Angwin et al., 2017; Barocas and Selbst, 2016; Chen et al., 2018), and the lack of using regional dialects in the training corpus of Natural Language Processing (NLP) algorithms (Jurgens et al., 2017).

One of the most researched areas of machine bias is related to NLP models—including what the main focus of this study is: bias in machine translation. Studies about machine translation (Prates et al., 2020; Cho et al., 2019) found that machine translation tools (e.g., Google Translate, Kakao translator, or Naver Papago) can exhibit gender bias and have a tendency to provide male defaults. Following the results of these previous studies, in this paper, we present a case study of Google Translate, a widely used machine translation tool. Previous studies of Google Translate (Prates et al., 2020; Cho et al., 2019) found gender bias in its operation. Gender bias in Google Translate manifests in the use of gender-based pronouns (i.e., "he" and "she"), because in some gender-based languages (e.g., in English) the translator—necessarily—provides either feminine or masculine translations to originally gender-neutral words. For example, English is a gender-based language having a male, a female, and a neutral version of the singular third-person pronoun, i.e., "he," "she," and "it"; while gender-neutral languages, such as Hungarian, do not have gender-based pronouns. Studies investigating the gender bias of Google Translate analyzed the English translations of sentences written in a wide variety of gender-neutral languages, including Hungarian, Korean, and Turkish. They examined the translations of sentences that included job positions or adjectives, like "he/she is a doctor" and "he/she is happy." Prates et al. (2020) analyzed the English translations of sentences containing gender-neutral pronouns, like "ő egy orvos" ("he/she is a doctor" in Hungarian)—where "orvos" means "doctor"; and "ő" is a gender-neutral singular third-person pronoun. In these sentences, the personal pronoun of the source language can be either translated into "he" or "she." Since, without any context, "he" and "she" are both correct translations of the singular third-person pronoun "ő," Google Translate's algorithm must choose between them and provide one correct, gender-based answer for the query. It creates the possibility of gender stereotypes appearing in the algorithm. **Figure 1** shows examples of translating gender-neutral pronouns into gender-based pronouns.

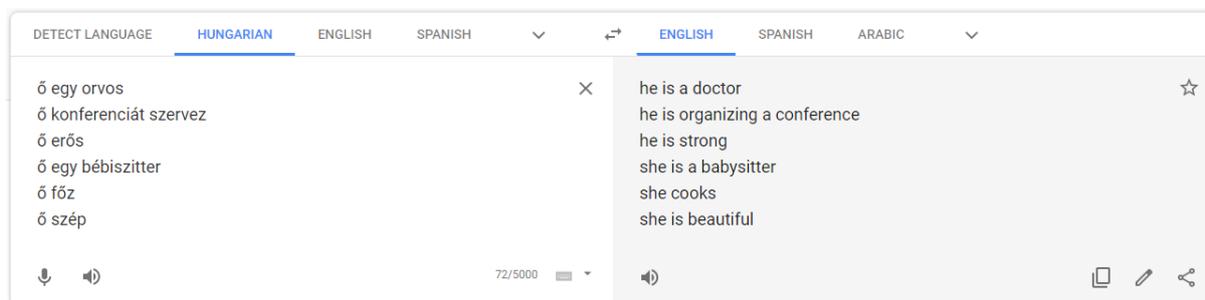

**Figure 1:** Examples of translating gender-neutral pronouns into gender-based pronouns .[1]

In this paper, we present a case study on job-related gender bias of Google Translate. In addition to the results of the specific case, we also give a methodological approach which can be generally applicable to investigate machine bias. This paper presents a systematic study of job-related gender bias in machine translation. As former studies captured the complex notion of bias with a single indicator (see, Prates et al., 2020), we aimed at providing a more complete and sociologically more grounded quantification of the phenomenon. The case study aims to measure the extent of gender bias appearing in the Hungarian–English translation of sentences including the names of occupations. We examine the translation of sentences, like "ő egy orvos" ("he/she is a doctor") or "ő egy mérnök" ("he/she is an engineer"). Analyzing these translations, we measure the extent of gender bias by comparing Google Translate's results to an optimal machine translator. In this paper, we present three types of optimal translators: (1) one that is based on the percentage of male and female workers in each occupation in Hungary according to the source language, (2) one that is based on the percentage of male and female workers in each occupation in the USA according to the target language, and (3) another that is based on whether people find a particular occupation feminine or masculine. In each of these cases, we contrast the Translate's results with the "real world." In the first two cases, real-world external data come from administrative data on social structure, while in the third case, we used self-reported survey data. The optimal translator gives results that fit best to the given external data.

---

[1] Source: Google Translate [Screenshot]:
https://translate.google.hu/?hl=en&tab=wT#view=home&op=translate&sl=hu&tl=en&text=%C5%91%20egy%20orvos%0A%C5%91%20konferenci%C3%A1t%20szervez%0A%C5%91%20er%C5%91s%0A%C5%91%20egy%20b%C3%A9biszitter%0A%C5%91%20f%C5%91z%0A%C5%91%20sz%C3%A9p (accessed 7 August 2020). Google and Google Translate are trademarks of Google LLC and this paper is not endorsed by or affiliated with Google in any way.

As growing complexity of sentences may introduce changes in bias, our paper also includes an investigation of sentences containing both occupations and adjectives that characterize them. We analyzed the English translation of Hungarian sentences, such as "ő egy jó orvos," "ő egy nagyon jó orvos," "ő egy rossz orvos," and "ő egy nagyon rossz orvos" ("he/she is a good doctor," "he/she is a very good doctor," "he/she is a bad doctor," and "he/she is a very bad doctor"). To our best knowledge, this is the first study to examine the role of occupations and adjectives combined in machine translation.

Before describing the case study on Google Translate, it is important to mention that for some gender-neutral languages, Google Translate has been serving both feminine and masculine translations since 2018 (Kuczmarski, 2018). This option is called "gender-specific translation" and it has been extended to Hungarian while preparing this study. Now it provides both feminine and masculine translations for a single sentence containing a Hungarian gender-neutral word. It translates "ő" into "he" and "she" as shown in **Figure 2**.

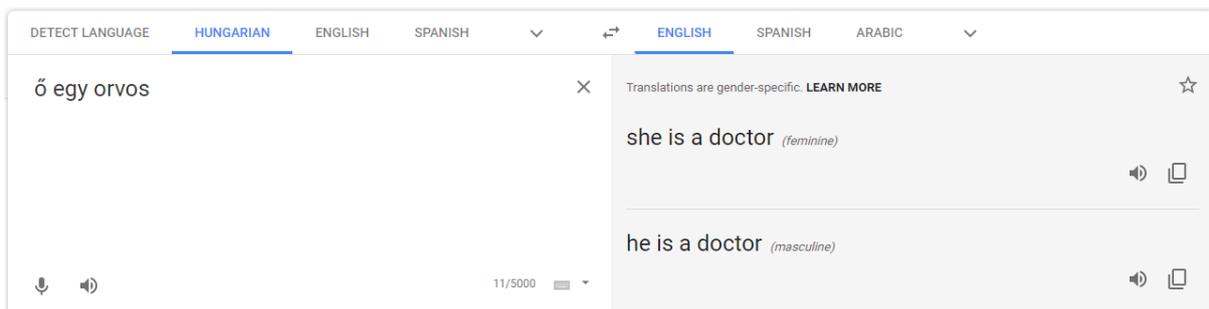

**Figure 2:** Example for gender specific translation.[2]

However, the analysis of Google Translate's results remains to be relevant for several reasons. First, the gender-specific translation of gender-neutral words is only provided for single sentences. When users want to translate multiple sentences (see, **Figure 1**) or they use Google Translate's function that translates whole documents, it still provides gender-based results. Secondly, gender-specific translations are only implemented in English translations, at the time, it is not available for Hungarian–Spanish or Hungarian–French translations or other language pairs. English often serves as an intermediary language between other languages (Prates et al., 2020), suggesting that biases in Hungarian–English translations may affect Hungarian–Spanish and Hungarian–French translations. Therefore, examining Hungarian–English language pairs remains to be relevant. Thirdly, the machine learning model used by Google Translate (called word embedding) is a widely used NLP model also utilized by Netflix,

---

[2] Source: Google Translate [Screenshot]:
https://translate.google.hu/?hl=en&tab=wT#view=home&op=translate&sl=hu&tl=en&text=%C5%91%20egy%20orvos (accessed August 7 2020). Google Translate is a trademark of Google LLC.

Spotify, and Google Search (Olson, 2018). Thus, the results of investigating gender bias in Google Translate can be relevant to investigate gender bias in other NLP models using word embedding. In this paper, we provide a methodology that can be a fruitful approach for examining the extent of bias in other machine learning tools.

## Previous Studies

Recently, the most effective machine translation algorithms use a deep learning-based model called word embedding (Laki, 2018). Several researchers (Bolukbasi et al., 2016; Olson, 2018) have observed that word embedding models tend to adopt gender biases, as they are sensitive to gender stereotypes present in data and society. The reason behind this is that word embedding models are based on the co-occurrence of words and phrases of a text corpus. Word embedding models represent words and phrases as vectors in a vector space. The more co-occurring the words are and the more similar their semantic meaning is, the closer they are in the vector space (Bolukbasi et al., 2016; Laki, 2018). For instance, the words "woman" and "mother" are expected to be close, since their semantic meaning is close. (For the same reason, similar vector proximity is expected from words "man" and "father.") Meanwhile, it raises questions when words like "doctor" or "strong" are connected to "man"; and "receptionist" or "beautiful" are connected to "woman," because, without context, those words could be equally related to "men" and "women." However, if the word "doctor" appears frequently in the context of "man" in the corpus the model is trained on, the model will connect "doctor" to "men" (Olson, 2018). That is why word embedding models tend to reinforce and recreate gender stereotypes existent in society—the same way as tools do while implementing word embedding models such as machine translation algorithms. Since 2016, Google Translate also uses word embeddings (Wu et al., 2016), which explains why it was found to exhibit gender bias. It is trained on a text corpus containing gender stereotypes—and, using word embeddings, the algorithm learns these stereotypes and creates gender asymmetries in the results. To resolve this problem, Google Translate found the solution to provide both feminine and masculine translations for gender-neutral words (called gender-specific translations). It is worth mentioning that there have been other attempts to debias word embedding systems, which were proved to be successful (Bolukbasi et al., 2016).

Previous studies about gender bias in Google Translate aimed to find a method to define gender bias in machine translation and identify it in Google Translate in case it is present in the translator. To our knowledge, Cho at al. (2019) and Prates at al. (2020) were the first ones to investigate gender bias in Google Translate from a social science perspective. Cho et al. (2019) compared Google Translate's results to two other translators (Kakao translator and Naver Papago) measuring the difference between their performance. Prates et al. (2020) compared the English translations of 12 gender-neutral languages to occupational statistics of the U.S., which allowed them to measure the gender bias of English translations in general. From a methodological point of view, Prates et al. captured the complex notion of bias with a single indicator (difference between translated results and current occupational statistics in the target language country). As compared to this study, we aimed to apply a broader approach to detecting bias and present a quantitative approach generalizable to any machine bias in general.

## Methods

### Defining Gender Bias

To measure the extent of gender bias in Google Translate, first, we need to conceptualize gender bias in machine translation. Bias, in general, refers to the phenomenon when our observations systematically deviate from an existing or theoretical benchmark (Freedman et al., 2005). Some people argue that the theoretical benchmark for machine translation is using female and male pronouns equally (Cho et al., 2019). However, deviating from this 50:50 ratio does not necessarily mean the algorithm is biased. It only reflects the social world, as the distribution of male and female employees in most occupations deviates from 50:50. Therefore, we define machine bias as an amplification of existing occupational gender segregation. As a benchmark, we define an "optimal" machine translator as one that simply reflects the real-life distribution of men and women in each occupation. For example, if an occupation is conducted by more women than men, our optimal translator translates a sentence containing the occupation with a female pronoun (i.e., "she"), the same logic applies for occupations that are conducted by more men than women. Prates et al. (2020) followed a similar approach in their study of Google Translate. They compared the translations of occupations with the distribution of men and women in those occupations as registered in the occupational statistics of the U.S. Bureau of Labor Statistics (BLS). They compared the translations only to U.S. labor force statistics which describes the society of the target language (i.e., English), however, a comparison to the society of the source language could be justified as well. The gender bias exhibited by Google Translate is the result of gender stereotypes apparent in the text corpus the algorithm was trained on. Those texts may have been written originally in the target language or the source language. In the case of Hungarian–English translations, the original texts may have been written either in Hungarian or in English, containing gender stereotypes and biases of either the Hungarian or an English-speaking society. Google Translate was originally trained on texts published by the United Nations and the European Parliament (Prates et al. 2019), which indicates that probably most of the texts in its training corpus were written in English. This would justify the comparison to U.S. statistics. However, since 2014, it also relies on data obtained from users (Prates et al., 2019; Kelman, 2014), which makes it reasonable to compare Google Translate's results to the society of the source language as well. Therefore, in this study, we compare the translation of occupations to both Hungarian and U.S. statistics.

To compare the translations to the male-to-female ratio of each occupation in Hungary, we used the 2011 Hungarian census data (Hungarian Population Census, 2011), which includes information about the distribution of men and women in 485 occupational categories. To compare the translations with the male-to-female ratio of each occupation in the U.S., we used data based on the Current Population Survey conducted in 2011—retrieved from the Bureau of Labor Statistics (2011). Unfortunately, the latter data had some weaknesses from our point of view. First of all, the BLS did not report the proportion of men and women in occupational categories with less than 50,000 employees, thus, we had to eliminate some U.S. occupational categories from our analysis, which means our comparison included only 239 U.S. occupational categories. Secondly, the BLS did not provide any data on military occupations, thus, we could only compare the translations of military occupations to Hungarian statistics. Lastly, the occupational categories of the U.S. statistics provided by the BLS have a different structure than the occupational categories of the Hungarian census. The BLS uses the SOC (Standard Occupational Classification) system; while the Hungarian census uses the so-called FEOR system. To compare the gender bias according to Hungarian statistics with the bias according to U.S. statistics, we had to pair the Hungarian FEOR occupational categories with the American SOC occupational categories. We used the International Standard Classification of Occupations (ISCO) system to create a crosswalk between the FEOR categories and the SOC categories (Bureau of Labor Statistics, 2015; Hungarian Central Statistical Office, n.d.[3]).

The distribution of male and female employees is not the only factor determining the gender asymmetries of the translations of occupations. The gender stereotypes and bias present in the training data of Google Translate are also influenced by the way society thinks and writes

---

[3] There are two sources from the Hungarian Central Statistical Office with no date. The source referenced here: https://www.ksh.hu/docs/osztalyozasok/feor/fordkulcs_feor_isco_hu.pdf (accessed 9 November 2020)

about those occupations. Earlier, we defined an optimal translator that follows the real-life male-to-female ratio of occupations. Following an alternative definition for "real-life," we also defined an optimal translator that follows our society's views on occupations and gender based on whether people find an occupation feminine or masculine. One of the most cited studies about gender stereotypes of occupations is Shinar's study written in 1976, in which participants had to rate occupations as masculine, feminine, or neutral on a scale from 1 to 8. Although several studies have been published based on Shinar's research (Beggs and Doolittle, 1993; Couch and Sigler, 2001), to our knowledge, there has not been any representative survey conducted recently neither in Hungary nor in the U.S that aimed to measure the gender stereotypes of occupations. Consequently, we were not able to compare the translations to any data or survey results indicating what occupations are considered feminine and masculine in the USA. On the other hand, to be able to conduct the same comparison of translations with Hungarian data, we designed a survey based on Shinar's study. The questionnaire measured gender stereotypes of occupations on a Likert scale. Respondents had to answer the question: "On a scale from 1 to 6, where 1 means it is very masculine and 6 means it is very feminine, how much do you consider the occupations below to be masculine or feminine?" (The original question was written in Hungarian for the Hungarian participants). The survey was part of an online omnibus survey with a sample size of 1,000 participants representing the Hungarian adult population. In total, one hundred occupations were included in the survey. To avoid bias coming from respondent fatigue, we randomly divided the occupations into five groups. 200 participants were asked to evaluate each group of occupations by labeling the occupations with a value from the masculine–feminine continuum (1–6). Thus, each participant had to label 20 occupations. The order of the 20 occupations was also randomized in order to reduce response bias.

When comparing the survey responses with the translations, we need a measure of masculinity/femininity for each occupation. As the original Likert-scale does not give such a score, we transformed it by keeping the distances between the original answer values unchanged. Responses 1 and 6 ("very masculine" and "very feminine") were transformed to 2.5, responses 2 and 5 ("masculine" and "feminine") were transformed to 1.5, and responses 3 and 4 ("somewhat masculine" and "somewhat feminine") got a value of 0.5. Masculinity score of an occupation emerges as a sum of values for responses (1–3) divided by the sum of values for all responses (1–6); while femininity score of an occupation emerges as a sum of values for responses (4–6) divided by the sum of values for all responses (1–6). An example for the calculation is shown in **Tables 1** and **2**: masculinity of the occupation "carpenter" is (425+18+3.5)/454=98%. Over the above, the survey included an additional question to assess the basis on which respondents decided whether they considered an occupation to be masculine or feminine.

|  | 1 – very masculine | 2 | 3 | 4 | 5 | 6 – very feminine | Total |
|---|---|---|---|---|---|---|---|
| carpenter | 170 | 12 | 7 | 3 | 4 | 0 | 200 |

**Table 1:** Frequency of the answers given for the question regarding the occupation "carpenter" → "On a scale from 1 to 6, where 1 means it is very masculine and 6 means it is very feminine, how much do you consider the occupations below to be masculine or feminine?"

|  | 1 – very masculine | 2 | 3 | 4 | 5 | 6 – very feminine | Total | masculinity score (% of total) | femininity score (% of total) |
|---|---|---|---|---|---|---|---|---|---|
| carpenter | 425 | 18 | 3.5 | 1.5 | 6 | 0 | 454 | 98% | 2% |

**Table 2:** Sum of values for answers given for the question regarding the occupation "carpenter" → "On a scale from 1 to 6, where 1 means it is very masculine and 6 means it is very feminine, how much

do you consider the occupations below to be masculine or feminine?" & the calculated masculinity/femininity scores.

**Measuring Gender Bias**

When evaluating the bias of a translator, we have to compare the functioning of the translator to the "reality" by using some real-world data. Following the above considerations, we compared results of Google Translate to three kinds of external real-world data: (1) the proportion of male and female workers of each occupation in Hungary, (2) the proportion of male and female workers of each occupation in the USA, and (3) the masculinity/femininity score of occupations. However, such a raw comparison is not fair: results of a translator cannot fit exactly to the external data. This is an issue former studies (see, Prates et al., 2020) failed to consider. To create a fair comparison, we defined an "optimal" translator that fits as well to the external data as possible.

In the following, we explain how to create an optimal translator that best fits to the male-to-female ratio of employees, and how to measure gender bias of a particular translator by comparing it to the optimal one. Supposing that 60% of the employees of occupation "A" are women and 40% are men, an optimal translator that strictly follows this ratio translates occupation "A" with "she" in 60% of the time and with "he" in 40% of the time. We call this translator an *optimal probabilistic translator,* because it translates a given occupation into "he" and "she" randomly with probabilities equaling its male-to-female ratio. Google Translate, on the other hand, is not a probabilistic translator but a deterministic one. An *optimal deterministic translator* translates occupation "A" with the pronoun "she," because it is conducted by more women than men. However, this translation does not represent 40% of the employees of occupation "A" who are men, so the translation gets 40 error points. The error of an optimal deterministic translator is called *optimal error* ($E_o$). "Optimal" because this is the smallest error achievable by a deterministic translator. A fair evaluation of the performance of Google Translate is based on a comparison to this optimal error. If Google Translate relates occupation "A" with "she," the Translate's error ($E_t$) will also be 40 error points, that is, it works in an optimal way. Meanwhile, if Google Translate relates occupation "A" with the pronoun "he," its error will be 60 error points since it does not represent 60% of the employees, who are women. The 20-point difference can be benchmarked against the optimal error. In the example, the comparison gives 20/40=0.5, that is, the translator is 50% more biased than an optimal deterministic translator. Formally, the extent of gender bias (B) in the Translate is:

$$B = \frac{E_t - E_o}{E_o}$$

If Google Translate translates a sentence with the adequate pronoun, its error will equal the error of an optimal deterministic translator and the extent of gender bias will be 0. Thus, if there are more female employees in a given occupation than male employees and the translator translates it with a female pronoun accordingly, we consider the translation to be unbiased. Accordingly, if the translator translates the sentence with an inadequate pronoun, we consider it to be biased. The score measuring the extent of gender bias can be any positive number.

A similar approach and formula were used to compare Google Translate's results to an optimal translator that is based on society's view on gender and occupations. If we suppose that occupation "A" is seen as rather feminine by 60% of the participants of the survey, and Google Translate relates it with the inadequate pronoun "he", Google Translate's error will be 60 error points and it will be evaluated to be 50% more biased than an optimal translator based on society's opinion.

Some examples are shown in **Table 3** displaying the score measuring the extent of gender bias when defined in relation to FEOR occupational statistics. As can be seen, for example, the strongly female-dominated "statistician" is translated with the pronoun "he."

There are some occupational categories, like "Dancers and Choreographers" which contain multiple occupations. Since we only had data about the male-to-female ratio of the occupational categories—and not individual occupations—, the gender bias score of occupational categories containing multiple occupations was calculated by averaging the bias over all occupations within the category.

| occupation | FEOR category | pronoun | female employees (%) | male employees (%) | gender bias score of the occupation | gender bias score of the category |
|---|---|---|---|---|---|---|
| statistician | Statisticians | he | 73 | 27 | 1.7 | 1.7 |
| dancer | Dancers and Choreographers | she | 58 | 42 | 0 | 0.2 |
| choreographer | Dancers and Choreographers | he | 58 | 42 | 0.4 | |

**Table 3:** The calculation of the gender bias score of occupational categories.

The scores measuring gender bias were calculated by comparing the translation of every occupation to occupational statistics (FEOR, SOC) and attitude survey results. Besides analyzing the extent of gender bias of occupations, we created larger occupational groups, so that we could examine the average gender bias arising in different occupational fields. We specified 18 larger groups of occupations with the help of the beforementioned FEOR, SOC, and ISCO occupational categorizations. We aimed to create groups that include similar occupations, however, are not too broad, as too broad grouping may cover up distortions in the case of certain groups. Number of occupations per groups is shown in **Table 3**. Since the structure of FEOR and SOC categories differ and the BLS did not provide any information about the male-to-female ratio of several SOC categories, the number of occupations a group involves may differ in the case of Hungarian and U.S. occupational categories.

| Employment sectors | number of occupations according to FEOR categories | number of occupations according to SOC categories |
|---|---|---|
| **Managers** | 31 | 19 |
| **Science and Engineering** | 69 | 28 |
| **Social Science** | 13 | 6 |
| **Healthcare** | 28 | 23 |
| **Culture, Arts, and Sports** | 27 | 13 |
| **Education** | 14 | 8 |
| **Business and Finance** | 18 | 15 |
| **Legal Occupations** | 5 | 3 |
| **Office and Administrative Occupations** | 27 | 20 |
| **Building Industry** | 20 | 14 |
| **Crafts and Light Industry** | 23 | 6 |
| **Industry, other** | 18 | 7 |
| **Service** | 46 | 37 |

| Sales | 14 | 13 |
|---|---|---|
| Agriculture | 15 | 4 |
| Machine Operators and Assemblers | 38 | 15 |
| Transportation | 15 | 8 |
| Military | 3 | 0 |

**Table 4:** Number of occupations and occupational categories within employment sectors.

By grouping occupations, we were able to measure the extent of gender bias in different employment sectors, such as Healthcare or Education. The extent of gender bias in a given employment sector equals the weighted average of gender bias in occupations within the group, with weights equaling the number of people in the given occupation. Since the ratio of female- and male-dominated occupations vary among these employment sectors, it would not be fair to compare the average bias score of the male-dominated and the female-dominated sectors. To resolve this problem, we analyzed the bias of the female- and male-dominated occupations of a sector separately. Analyzing the average gender bias of male- and female-dominated occupations in different employment sectors helped us to explore which sectors are associated with gender stereotypes the most.

## A Step Forward: When Adjectives Characterize the Occupations

As the growing complexity of sentences may introduce changes in bias, we, as a complementary study to our research about occupations, examined sentences containing both occupations and adjectives. This enabled us to explore the role of occupations' names and adjectives in determining the translation of a sentence. There have been previous studies that investigated the gender bias of sentences containing adjectives (Cho et al., 2019; Prates et al., 2020). They analyzed the translation of sentences, like "he/she is happy" or "he/she is sad." However, to our best knowledge, this is the first case study that examines the translation of sentences that contain both occupations and adjectives. We chose two adjectives, "good" and "bad," as attributives and created sentences that followed this structure: "ő egy jó orvos," "ő egy nagyon jó orvos," "ő egy rossz orvos," "ő egy nagyon rossz orvos" ("he/she is a good doctor," "he/she is a very good doctor," "he/she is a bad doctor," "he/she is a very bad doctor"). As there is no existing benchmark for determining the gender of "good" and "bad doctors", we could not compare the results of these sentences to an optimal translator that does not exhibit gender bias. Instead, we examined how adjectives change the gender of the pronoun used in the original sentences that contained only occupations. The comparison allowed us to measure the role of names of occupations and adjectives in determining the gender of the pronoun in the translation.

## Data

To analyze the functioning of Google Translate, we had to create a list of occupations and had to put them into sentences that followed the structure of "ő egy. . ." (meaning "he/she is a. . ."). To create a list of occupations, we used the Hungarian Standard Classification of Occupations of the Hungarian Central Statistical Office (FEOR classification, Hungarian Central Statistical Office, n.d.[4]). Using the FEOR classification for making a list of occupations proved to be effective for several reasons. First of all, the source language of our translations was Hungarian, so it was obvious to use a Hungarian classification. Secondly, using the FEOR system allowed us to compare the translations to the Hungarian census-based occupational statistics that are given also according to the FEOR system. Thus, we had data on the male-to-female ratio of every FEOR occupational category. Thirdly, as mentioned earlier, we created a crosswalk between the FEOR and the SOC systems, which allowed us also to assign the American male-to-female ratio to our list of occupations There were some cases when the official SOC category associated with a FEOR category did not describe a given occupation accurately. In those cases, we changed the SOC category to a more appropriate one, using the occupation code finder developed by the U.S. Department of Labor (National Center for O*NET Development, 2020).

---

[4] There are two sources from the Hungarian Central Statistical Office with no date. The source referenced here: https://www.ksh.hu/docs/szolgaltatasok/hun/feor08/feorlista.html (accessed 9 November 2020)

Some of the occupational categories of the FEOR system describe one job position, while others contain multiple occupations. For the latter, it was necessary to divide the category into several subcategories. When selecting the occupations, we strived for selecting occupations that are well-known and, thus, more likely appearing in the training corpus of Google Translate. Therefore, occupations that correspond to special positions mentioned only in the FEOR system were not included in the final list of occupations. Additionally, some job positions in Hungarian define the gender of the person holding the position, e.g., "védőnő" refers to a female healthcare practitioner. Other occupations have different forms for male and female workers, like "színész" and "színésznő" (meaning "actor" and "actress"). Although "színész" can refer to both female and male actors, it is somewhat more common to use it for male actors. These occupations had to be excluded from the list, because they determine the gender of the pronoun in the translation. The same reasoning applies to religious occupations. Religious occupations determine (or at least are strongly related to) gender, and therefore, they are not included in our list. The final list includes 742 occupations. A few of them are shown in **Table 4** together with their FEOR and SOC categories.

| occupation | | FEOR category | SOC category |
|---|---|---|---|
| veterinarian | a | Állatorvos | Veterinarians |
| database designer | b | Adatbázis-tervező és -üzemeltető | Database designers and administrators |
| database operator | b | Adatbázis-tervező és -üzemeltető | Database designers and administrators |
| hairdresser | c | Fodrász | Hairdressers, Hairstylists, and Cosmetologists |
| barber | c | Fodrász | Barbers |

**Table 5:** Examples from the occupational list (translated into English) and their FEOR and SOC categories.
    a. The occupational category describes one occupation.
    b. The FEOR category contains multiple occupations.
    c. The FEOR category is associated with multiple SOC categories, thus, multiple occupations were assigned to the FEOR category.

The 100 occupations—included in the survey that aimed to measure the femininity and masculinity of these occupations—were chosen from the above list of 742 occupations. We selected occupations that are included in the FEOR categories as separate categories, so that we could compare the male-to-female ratio of occupational categories with survey results. In addition, we tried to include male- and female-dominated occupations equally.

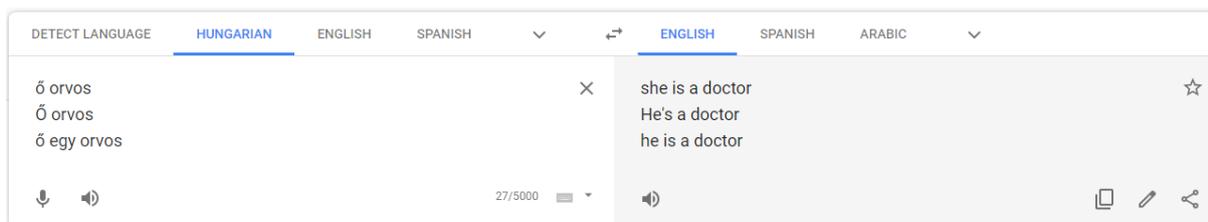

**Figure 3:** Different options for sentence structure alter the pronoun used in translation.[5]

Before analyzing the translation of sentences, we had to create sentences from the occupations. These sentences followed the structure of "ő egy orvos" ("he/she is a doctor"). It is important to mention that the sentences chosen to be translated can have different structures. They can begin with a capital letter ("Ő egy orvos") or follow the structure of "ő orvos" (the indefinite article "egy" meaning "a(n)" can be dropped from the Hungarian version of the sentence). These minor changes have an effect on the translations as shown in **Figure 3**. For our case study, we decided to use "ő egy orvos" because previous studies used the same structure (Prates et al., 2020). To create sentences with each occupation from our list, we used a code written in the programming language Python. These sentences were translated by using a special function of Google Translate with which it is possible to translate whole documents. At the time, the function translating documents provides only one gender-based pronoun for every sentence. It is important to mention that Google Translate's algorithm might have changed since we obtained the translations for this case study. All translations were retrieved in April 2020, hence, our results show the extent of gender bias of translations at that time.

All data collected and created for this case study can be accessed at: https://genderbiasdata.000webhostapp.com/

## Results

### Bias Defined in Relation to the Male-to-Female Ratio of Occupations

First, we present results on Google Translator's bias defined in relation to the male-to-female ratio of occupations according to the Hungarian census data. 35% of the occupations were translated with an inadequate pronoun, of which 77% were translated with "he" instead of "she." It suggests that although there is a bias against both genders, biased results against women are much more common. If the occupations were supposed to be translated with "she," Google Translate translated the occupation with "he" in 67% of the time. Meanwhile, if the correct pronoun was supposed to be "he," Google Translate used "she" in only 14% of the time. The probability of producing an incorrect translation for occupations that are female-dominated in Hungary is much greater than the probability of producing an incorrect translation for male-dominated occupations.

When the translations are compared to U.S. occupational statistics, a similar pattern appears. Google Translate translated the U.S. SOC occupational categories with an incorrect gender in 44% of the time, which is slightly greater than the error ratio of the translations compared to the Hungarian FEOR occupational categories. When the translations were incorrect compared to U.S. statistical data, Google Translate should have used a female pronoun in 71% of the time and a male pronoun in 29% of the time. 73% of the female-dominated and 22% of the male-dominated occupations were translated incorrectly. These results indicate a tendency similar to the bias measured against Hungarian occupational statistics.

---

[5] Source: Google Translate [Screenshot]:
https://translate.google.hu/?hl=en&tab=wT#view=home&op=translate&sl=hu&tl=en&text=%C5%91%20orvos%0A%C5%90%20orvos%0A%C5%91%20egy%20orvos (accessed 7 August 2020). Google Translate is a trademark of Google LLC.

We measured the extent of bias of Google Translate compared to an optimal deterministic translator, with a score ranging from 0 to any large positive number. Bias based on the Hungarian and the U.S. statistics showed a similar picture. Where we found gender bias compared to Hungarian statistics, the extent of bias ranged from 0.001 to 173.44 with a median of 1.01. Where we found gender bias compared to U.S. statistics, the bias score ranged from 0.01 to 88.91 with a median of 0.85.

Study results for employment sectors can be seen in **Figure 4** and **5**. Since the employment sectors have varying number of male- and female-dominated occupations, and employment sectors containing a lot of female-dominated occupations have a greater chance to get a higher average bias score, we analyzed male- and female-dominated occupations per sector separately. We plotted the bias of the employment sectors averaged over the female- and male-dominated occupations within the groups (with weights equaling the number of women/men in the given occupation).

**Figure 4** shows the average bias of sectors defined in relation to Hungarian statistics. Building Industry has the highest average bias score, which comes from a bias against male employees. All other sectors exhibit a bias against women, or a bias against men and women nearly equally. It is worth mentioning that none of the translations of Military occupations proved to be biased. **Figure 5** shows the average bias results defined in relation to U.S. statistics. Bias against men is noticeable in the building industry and service. Besides these employment sectors, in most sectors, the extent of gender bias against women is greater than or nearly equal to the gender bias against men. All in all, it can be concluded that the average gender bias, when defined in relation to either Hungarian or U.S. statistics, mainly derives from bias against female-dominated occupations with a few outlier sectors.

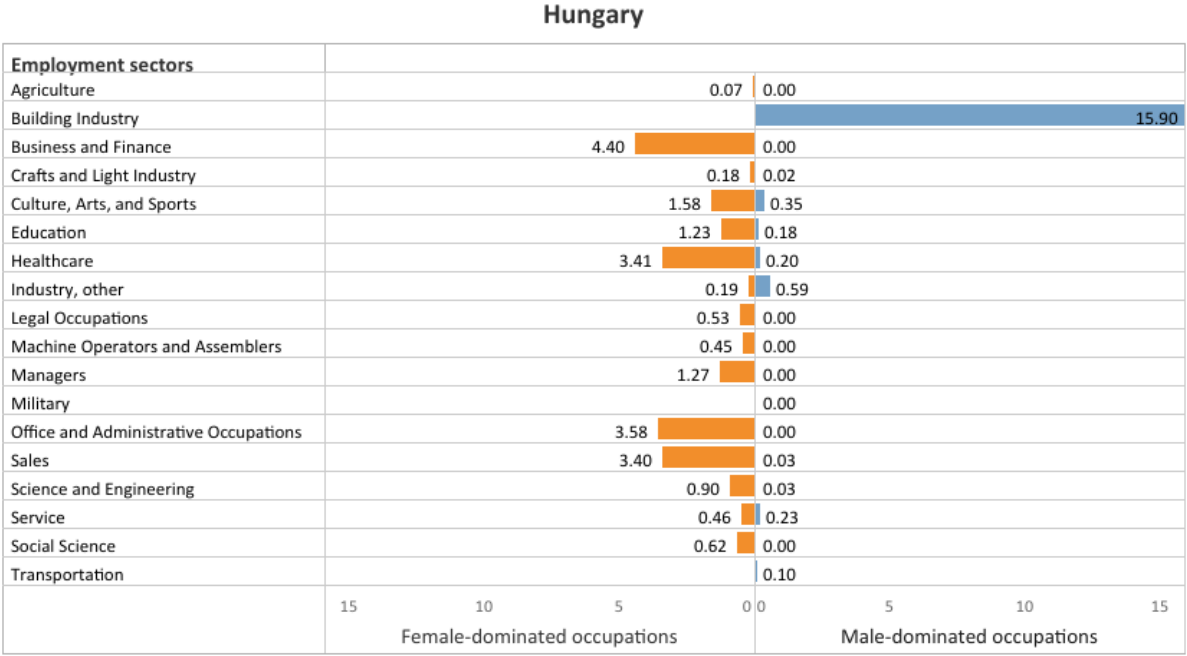

**Figure 4:** Weighted average of gender bias in female- and male-dominated occupations in Hungary by employment sector.

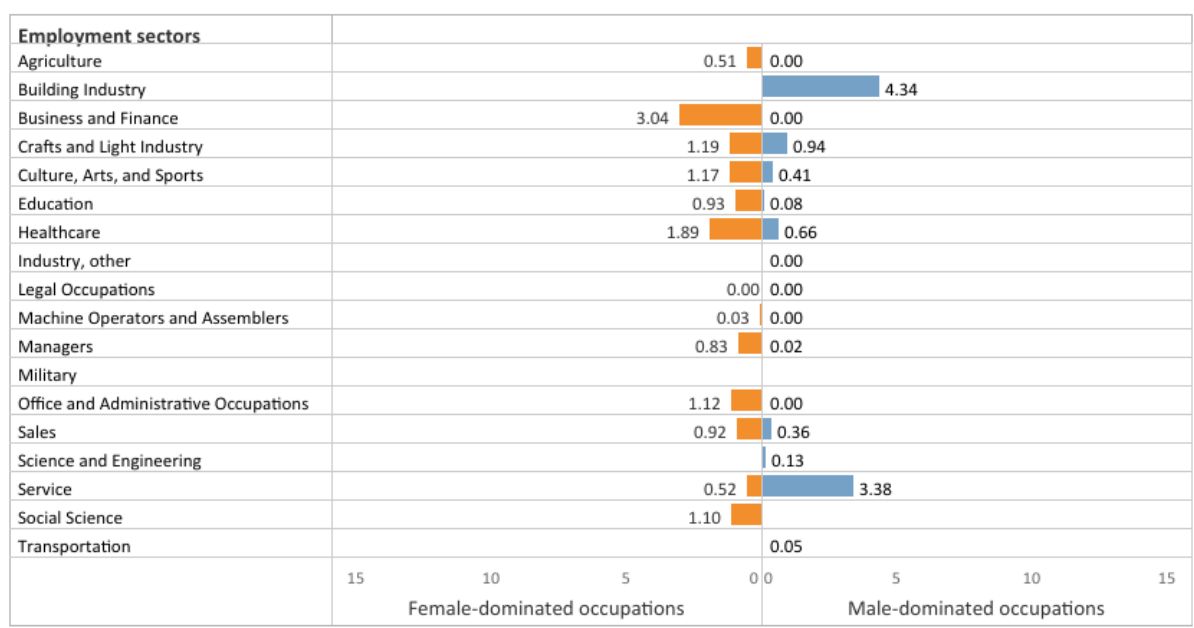

**Figure 5:** Weighted average of gender bias in female- and male-dominated occupations in the USA by employment sector.

**Bias Based on the Perceived Femininity and Masculinity of Occupations**

In this chapter, we present the results of measuring gender bias in Google Translate when defined in relation to society's opinion about the femininity and masculinity of occupations. **Figure 6** shows the femininity score of occupations (based on the percent of survey respondents who consider a given occupation feminine). It also gives information about how consistent the femininity score is with the percent of female employees of the given occupation. Occupations above/below the 45-degree diagonal have a higher/lower proportion of women than what could be expected from survey results. There is a moderate strong positive association between the perception of occupations and their real-life male-to-female ratio. There are a few outliers, like "shoemaker" or "administrative leader," as the figure shows.

      The discrepancy between the perception and real-life male-to-female ratio of those occupations can be assessed by using the additional survey question on the respondents' motivation. The question asked the reason, why the respondents decided whether they considered the occupations to be masculine or feminine, with four possible answers: (1) based on the proportion of men and women in the occupations, (2) based on personality traits related to the occupation, (3) based on physical abilities required for the occupation, and (4) based on other factors (an open-ended category). Only 20% of the respondents said that they based their answers on the proportion of men and women of occupations, while 30% of them made a decision based on personality traits related to the occupation, and 47% of them made a decision based on physical abilities required for the occupation. Some respondents mentioned that they took more than one aspect into account when they made a decision, and some had difficulty deciding, because they felt that "nowadays there are no longer jobs that are masculine or feminine" (quote from a respondent translated into English).

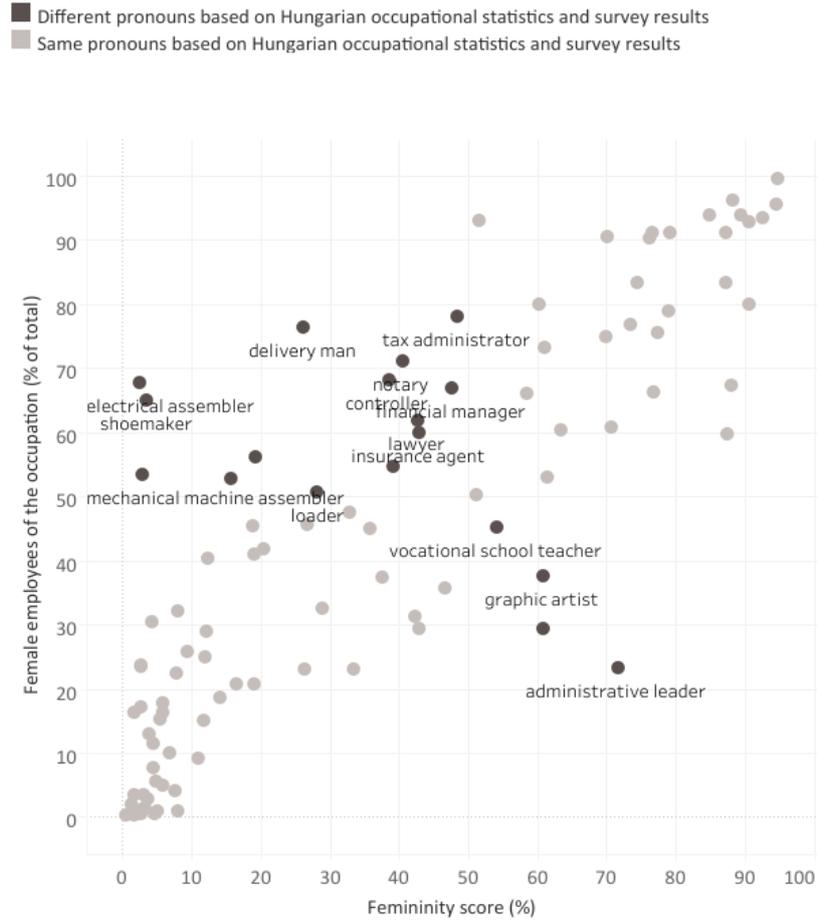

**Figure 6:** Scatter plot of the perceived femininity of the occupations in the survey and the proportion of their female employees according to Hungarian census data.

In the following, we analyze the extent of gender bias based on the survey. Google Translate results proved to be closer to the answers of the respondents than to the real-life male-to-female ratio of occupations. It translated only 20% of the occupations with a wrong pronoun. In comparison, the same value is 30% when compared to Hungarian occupational statistics. Out of 20 cases, where the algorithm was incorrect compared to survey results, the sentences were supposed to be translated with a female pronoun in 17 cases and with a male pronoun in only 3 cases, as shown in **Figure 7**. **Figure 7** also shows the gender bias score of the wrongly-translated 20 occupations.

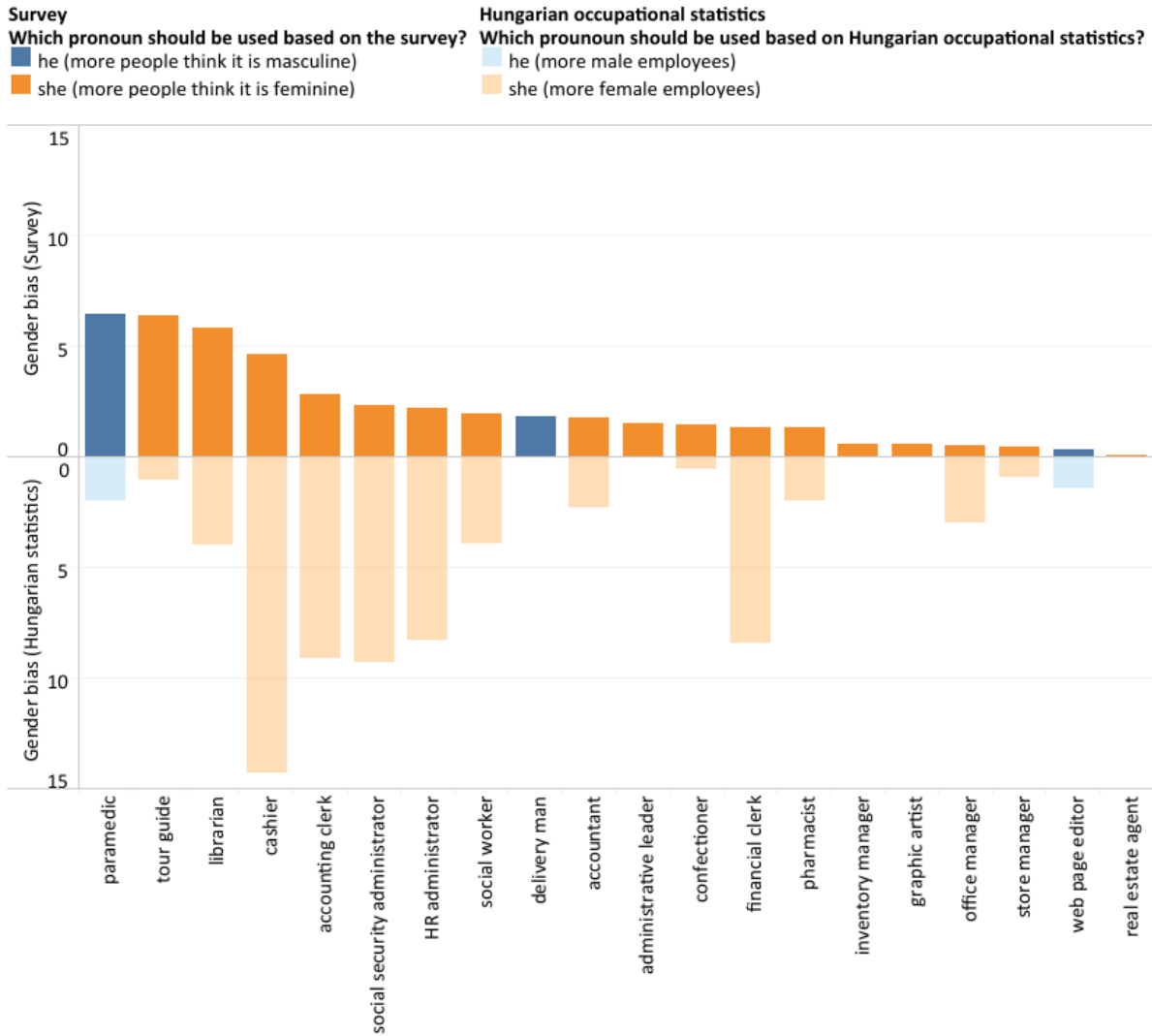

**Figure 7:** Bar chart of the 20 wrongly-translated occupations according to survey results showing the extent of their gender bias defined in relation both to the survey data and to Hungarian statistics.

If the sentences were supposed to be translated with a female pronoun according to the survey results, Google Translate translated the sentence with the wrong pronoun half the time. If the correct pronoun was supposed to be "he," it used "she" in only 4,5% of the time. Although the 100 occupations included in the survey are not a representative sample of the full list of occupations, the tendencies of gender bias based on survey results show a similar pattern to the tendencies of gender bias based on occupational statistics.

**The Effect of Adjectives in the Sentences to Be Translated**

In this chapter, we present our results of the complementary study of occupations and adjectives, where we examined sentences, like "he/she is a good doctor," "he/she is a very good doctor," "he/she is a bad doctor," "he/she is a very bad doctor." **Table 6** shows how the pronouns of the original sentences changed when adjectives were used. Although pronouns did not change in the majority of the sentences; when they did change, with one exception, they changed from "she" to "he." Nevertheless, adjectives altered the original pronoun of only 5–12% of the occupations, depending on the adjective used. This suggests that, in the examined sentences, occupations have a greater effect on the translations than adjectives.

| Changes compared to the original sentences (original sentence → sentence with an adjective) (% of total) | | | | | | |
|---|---|---|---|---|---|---|
| Adjective | she → she | he → he | she → he | he → she | did not change | changed |
| good | 16% | 79% | 5% | 0% | 95% | 5% |
| very good | 14% | 79% | 7% | 0.1% | 93% | 7% |
| bad | 10% | 79% | 11% | 0% | 89% | 11% |
| very bad | 9% | 79% | 12% | 0% | 88% | 12% |

**Table 6:** The proportion of pronouns changed due to the use of adjectives.

**Table 7** shows the proportion of male and female pronouns used in the sentences containing occupations and adjectives characterizing them. We found that 79% of the occupations were translated with a male pronoun and only 21% of them were translated with a female pronoun when only occupations were used. This tendency was even more asymmetrical for sentences where adjectives were used as well. When adjectives modified the original pronoun, the pronoun changed from "she" to "he" in almost every case as seen in **Table 6**.

| Adjective | she | he |
|---|---|---|
| NONE (only occupation) | 21% | 79% |
| good | 16% | 84% |
| very good | 14% | 86% |
| bad | 10% | 90% |
| very bad | 9% | 91% |

**Table 7:** The proportion of masculine and feminine translations.

## Discussion

Building on previous studies that found gender bias in Google Translate and other machine translation tools, we presented an approach in this paper for a systematic study of gender bias in machine translation algorithms. To give a complete picture, we tried to cover the fullest possible range of occupations, and measured machine bias both for particular occupations and also for larger employment sectors.

With the aim to define a fair measure for bias, we compared the translations not to the actual gender composition of occupations (as e.g., Prates et al., 2020) but to an optimal non-biased translator. For Hungarian–English translations of occupations, we established three benchmark criteria that an optimal Hungarian–English translator can be based on. The first two optimal translators aim to reflect the male-to-female ratio of occupations in Hungary and the USA, while the third optimal translator reflects the perceived femininity and masculinity of occupations. Comparing Google Translate's results to all three types of optimal translators, we confirmed the findings of previous studies that discovered gender bias in the algorithm (Cho et al., 2019; Prates et al., 2020). Comparing translations either to the Hungarian or the U.S. occupational statistics, we find a remarkable bias against both genders, but biased results against women are much more frequent. Also, in larger employment sectors, gender bias derives mainly from bias against women. We examined the average bias score of 18 employment sectors, e.g., Healthcare, Business and Finance, and Education. We found that in almost every sector, the average bias against women was bigger than the average bias against men—or they were nearly equal.

When comparing the translations to perceived masculinity/femininity of occupations instead of objective occupational statistics, we found the bias to be much lower, but still dominantly present against women. The result that Google Translate gives translations that are closer to our perception of occupations than to objective occupational statistics is plausible, as (1) they are perceptions and not actual social structures that are presented in the training corpus of Google Translate, and (2) changes in perceptions follow the actual structural changes with a delay.

We hypothesize that a more apparent bias against women can be explained by the use of male defaults. Google Translate uses hundreds of millions of texts collected from the Internet in order to determine its translations (Kuczmarski, 2018), but researches show that online texts contain more male-related words than female-related ones (Bolukbasi et al., 2016; Schiebinger, 2014). Consequently, male-related words are overrepresented in the training corpus of Google Translate, hence, it uses male pronouns more frequently than female pronouns. If the probability of using a male pronoun is higher, it will be likely to use a male pronoun even for occupations that are female-dominated. Naturally, machine translation algorithms are more complicated than the process described above, but the overrepresentation of male-related words in the dataset can cause a tendency where gender bias against women is more common compared to gender bias against men. Similarly to previous studies, our findings reinforce the assumption that male pronouns are more regular in Google Translate. The predominant use of male pronouns is more noticeable in sentences containing both occupations and adjectives (see, **Table 7**). This implies that a more complex sentence structure raises the probability of using a male pronoun—this theory, though, needs further proof and investigation.

The score created to measure the extent of gender bias proved to be efficient for quantifying gender bias in machine translation. Our methodology raises several possibilities for further research. Based on the formula of the gender bias score, follow-ups could be planned to investigate whether the structural changes in the job market cause changes in the translation of occupations. As another theoretical benchmark, the prestige scores associated with occupations could be also used to assess bias in translation. Finally, gender bias in other fields beyond occupations can be also quantified.

In this study, we discussed the issue of social bias found in machine learning models. The paper described a methodology to measure gender bias in machine translation through a case study of Google Translate. We think that by investigating bias in machine learning and finding tools to measure its extent, this paper may provide relevant findings to the ongoing discussion of the societal consequences of machine learning algorithms and may contribute to the mitigation of the resulting social risks.

## Acknowledgments

We thank the Inspira Group research company for generously incorporating our questions into their omnibus survey.

ELTE Eötvös Loránd University, Budapest, Hungary supported the proofreading of the paper.